\documentclass[letterpaper, 10pt, conference]{ieeeconf}
\IEEEoverridecommandlockouts

\usepackage{graphicx}
\usepackage{amsmath}
\usepackage{subcaption}
\usepackage{here}
\usepackage{tabularx}
\usepackage[table]{xcolor}
\usepackage{siunitx}

\begin{document}

\makeatletter

\title{\LARGE \bf
Soft-Jig: A Flexible Sensing Jig for Simultaneously Fixing and Estimating Orientation of Assembly Parts
}

\author{Tatsuya Sakuma, Takuya Kiyokawa, Jun Takamatsu, Takahiro Wada, and Tsukasa Ogasawara%
\thanks{All authors are with Division of Information Science, Nara Institute of Science and Technology (NAIST), Japan {\tt\footnotesize \{sakuma.tatsuya.sn1, kiyokawa.takuya, j-taka, t.wada, ogasawar\}@is.naist.jp}}%
}

\maketitle
\thispagestyle{empty}
\pagestyle{empty}

\begin{abstract}
For assembly tasks, it is essential to firmly fix target parts and to accurately estimate their poses.
Several rigid jigs for individual parts are frequently used in assembly factories to achieve precise and time-efficient product assembly.
However, providing customized jigs is time-consuming.
In this study, to address the lack of versatility in the shapes the jigs can be used for, we developed a flexible jig with a soft membrane including transparent beads and oil with a tuned refractive index.
The bead-based jamming transition was accomplished by discharging only oil enabling a part to be firmly fixed.
Because the two cameras under the jig are able to capture membrane shape changes, we proposed a sensing method to estimate the orientation of the part based on the behaviors of markers created on the jig's inner surface.
Through estimation experiments, the proposed system could estimate the orientation of a cylindrical object with a diameter larger than 50 mm and an RMSE of less than 3$^{\circ}$.
\end{abstract}

\section{Introduction}
\label{sec:introduction}

In many assembly operations, fixtures and jigs are designed to fit the shapes of a particular part. Such tools used to fix or support assembly parts are frequently used to firmly determine or control part orientations such as in insertion operations. However, the design and production of these tools require a large amount of time and high cost.
To achieve high-mix low-volume production in a short time at a low cost, we require a versatile assembly jig to replace the current custom-made jigs designed according to each product.

Kiyokawa \textit{et al.}~\cite{kiyokawa2020softjig} was the first attempt to use soft materials for a jig. They proposed a soft-jig made of a silicone membrane filled with only beads. This jig worked according to a jamming transition to fix parts, such as a jamming gripper~\cite{brown2010}, by vacuuming the air inside the jig.

If we use part-customized rigid jigs, the object orientation on the jig can be precisely determined because of the high rigidity of the surrounding planes that firmly constrain the object's surfaces. In contrast, if we use a soft-jig with a high degree of freedom in terms of the surface changes, the uncertainty of the jig-fixed object orientation is a crucial issue.

\begin{figure}[t]
  \centering
  \includegraphics[width=0.67\columnwidth]{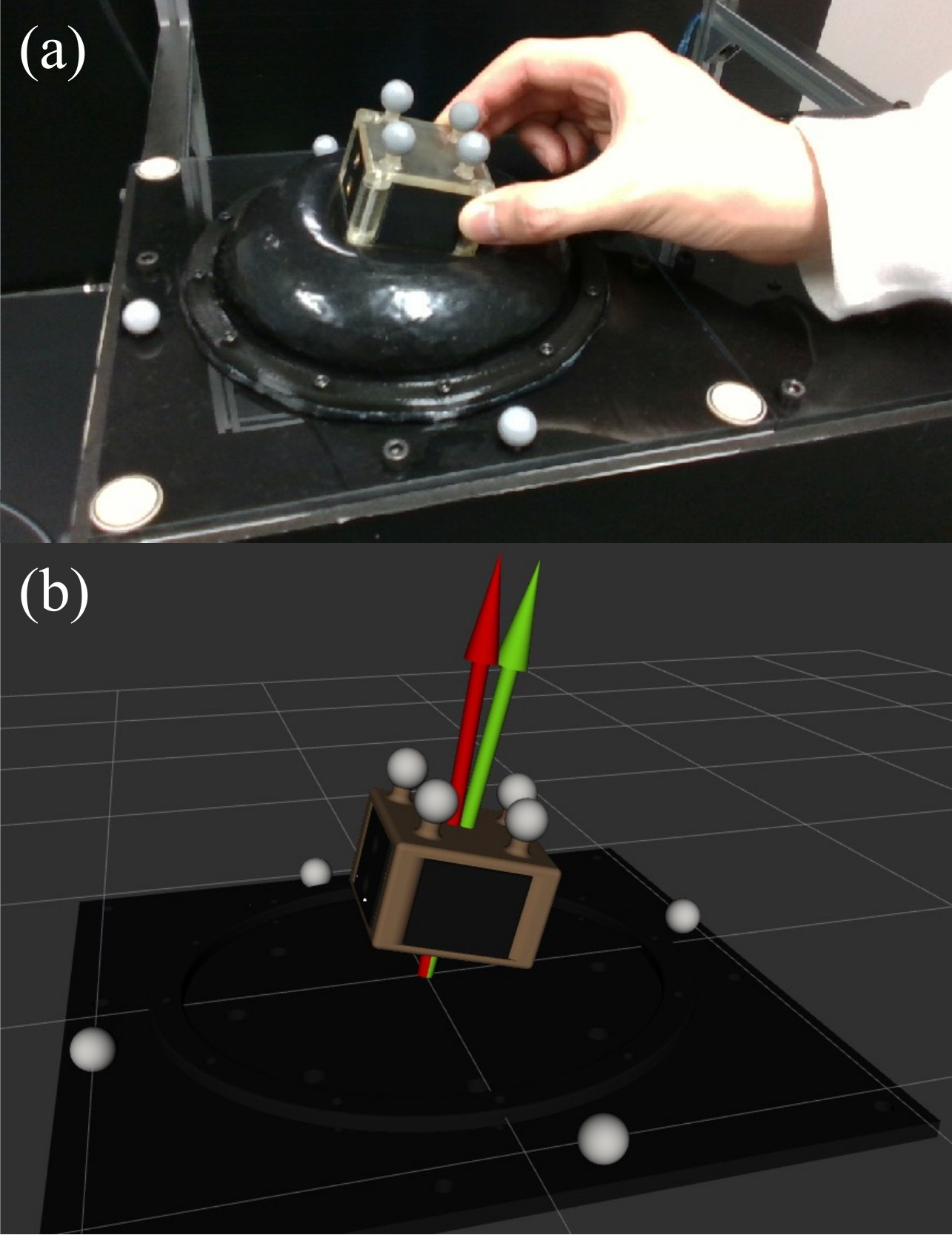}
  \caption{(a) Soft-jig with a thin flexible membrane filled with transparent beads and oil. (b) Principal normal vector of the target object (cube) estimated by the proposed soft-jig system (red arrow) and the normal vector measured using a motion capture system (green arrow).}
  \label{fig:overview}
\end{figure}

To reduce the uncertainty, the orientation of an object can be estimated using its image captured through an external camera.
Several state-of-the-art methods for 3D position and orientation estimation have achieved high accuracy~\cite{du2021}.
However, when the object is occluded by the gripper and jig, the orientation estimation deteriorates significantly.
Therefore, it is difficult for a robot to manipulate an object with high accuracy using only an external camera.

We attempted to reduce the uncertainty of the object orientation by using a sensing, flexible jig that acquires tactile information from the parts in contact with the jig.
Because tactile information is independent of the color and material of the object, it can be treated as highly reliable information.

Therefore, to estimate the orientation of a part while fixing the part, we used transparent beads and oil with a refractive index close to that of the beads. In this study, to recognize the markers on the inner surface of the jig, we replaced the air used for jamming grippers with transparent oil based on previous studies~\cite{sakuma2018}~\cite{sakuma2019}.
If the refractive indices of all the materials filled inside the jig are the same, when light travels straight through, the surface of the membrane can be captured optically by the camera.
To convert the captured image into tactile information that can be used by a robot, markers are painted on the inner surface of the membrane, and two cameras are installed inside the jig.

As part orientation needs to be estimated, we proposed to calculate a principal normal vector that shows the bottom surface normal of the target object. In our estimation method, we detect the markers on the membrane, estimate marker positions, calculate an approximate plane for the markers, and determine the plane normal as the principal normal vector, as shown in Fig.~\ref{fig:overview}.
The principal normal vector reduces the uncertainty of the orientation of the object on the jig and allows us to perform more precise assembly operations by re-grasping or changing the assembled object's orientation with the target orientation.

To evaluate the sensing ability of the proposed soft-jig, we conducted experiments on the principal normal vector estimation using three target objects in different orientations.

\begin{figure}[t]
  \centering
  \includegraphics[width=1.0\columnwidth]{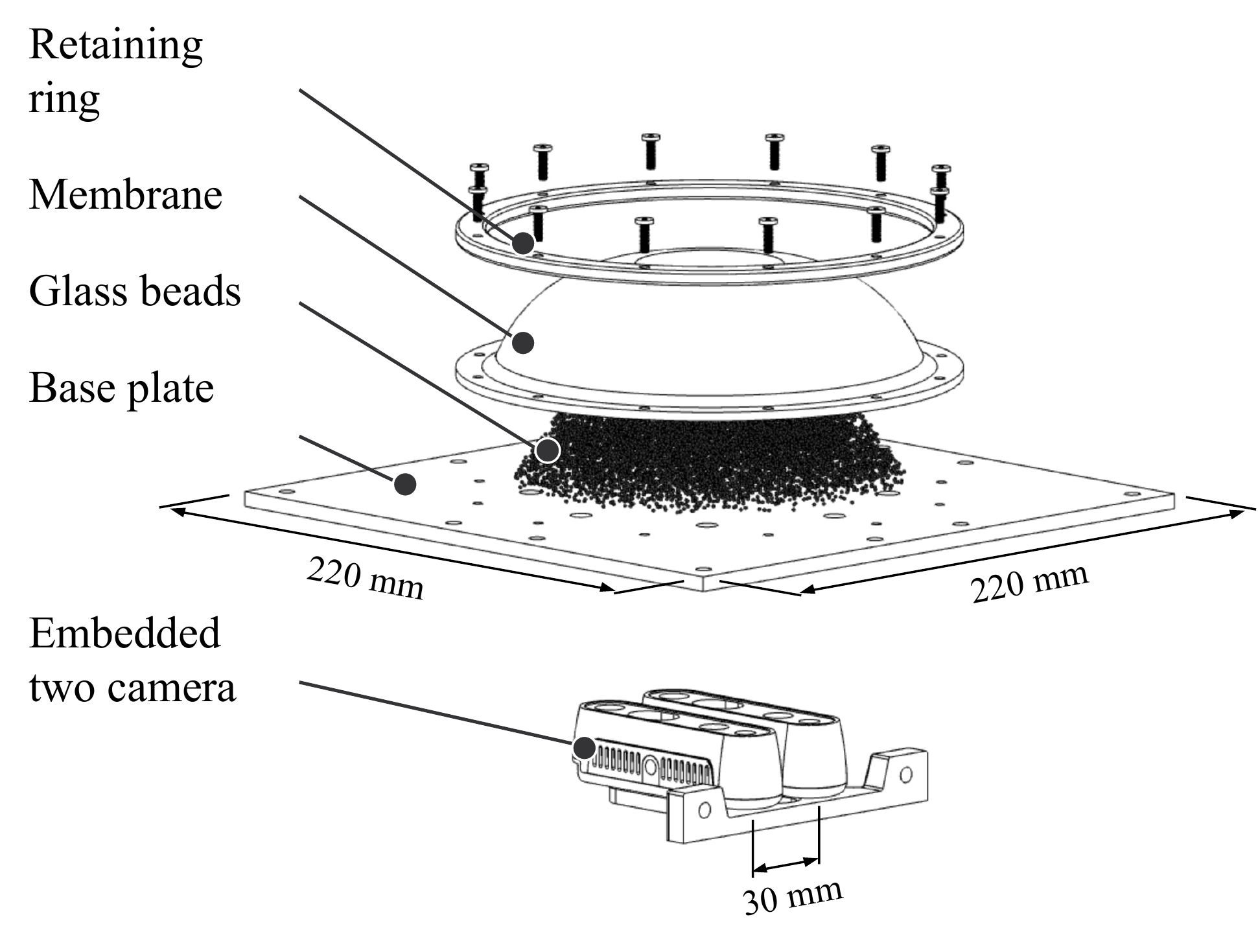}
  \caption{Design of the proposed soft-jig}
  \label{fig:design}
\end{figure}

\section{Related work}
\label{sec:related_work}

Flexible fixtures have been developed using a wide range of strategies.
Reconfigurable fixtures~\cite{lu2016}~\cite{muller2013} fix parts by rearranging or replacing the components. The ability to change in a fixed configuration provides flexibility.
Pin array fixtures~\cite{mo2019}~\cite{shi2021} use a shape-memorable mechanism to fix parts by constraining multiple points of the part with pins. The high degree of freedom in fixing using pins leads to flexibility.
Because these fixtures can fix different types of objects, they are suitable for high-mix low-volume production in terms of part-fixing; however, they do not aid in the uncertainty of the fixed object's orientation.
In addition, the proposed innovative fixture not only fixes parts using the jamming transition but can also perform sensing to acquire the principal normal vector simultaneously.

For sensors that use flexible materials to acquire the object shape on the soft jig, camera-based methods are attracting attention because of their high spatial resolution.
\textit{GelSight} is a sensor that acquires normal map using multiple light sources of different colors and photometric stereo~\cite{yuan2017}.
\textit{SoftBubble}~\cite{alspach2019},~\cite{kuppuswamy2020} relies on a depth camera to capture the shape of a balloon inflated with air and acquire the shape of the contact area.
Lin \textit{et al.}~\cite{lin2020} proposed a sensor that estimates the curvature of a flexible material using the subtractive color mixing principle.
In our previous work, we used a pinhole camera model to estimate the depth of a monocular camera using known-sized markers~\cite{sakuma2018,sakuma2019}.

Our proposed system performs the jamming transition to fix the object's orientation, and membrane shape acquisition using triangulation based on estimated marker positions and object orientation estimation for the object contacted or fixed.
A depth camera can be used for estimating the object's orientation, but it is difficult to acquire point cloud data using a commercial depth camera because the light passes through a transparent plate, oil, and beads to reach the inner surface of the membrane with attached markers.

\begin{figure}[t]
  \centering
  \includegraphics[width=0.7\columnwidth]{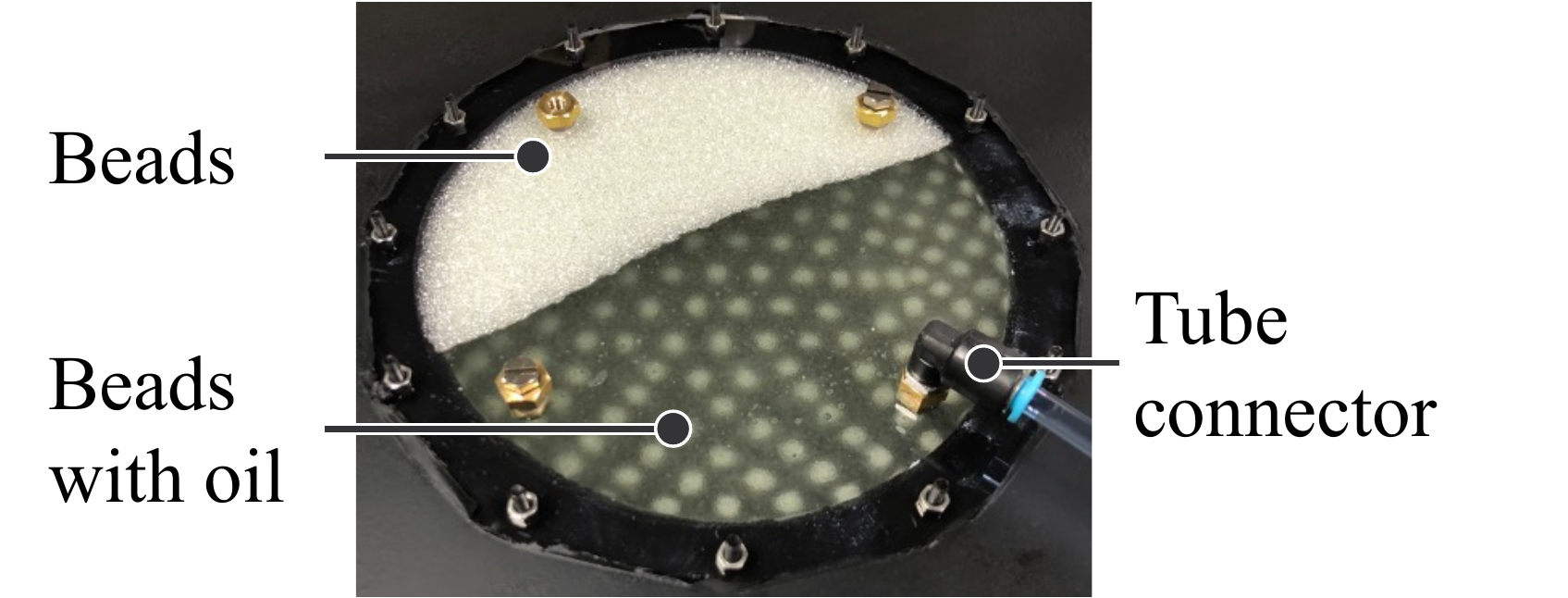}
  \caption{Appearance of soft-jig filled with the oil when viewed from the camera side. Even though the beads are packed tightly inside the membrane, the markers can be seen through the filling oil.}
  \label{fig:filling}
\end{figure}

\begin{figure}[t]
  \centering
  \includegraphics[width=0.8\columnwidth]{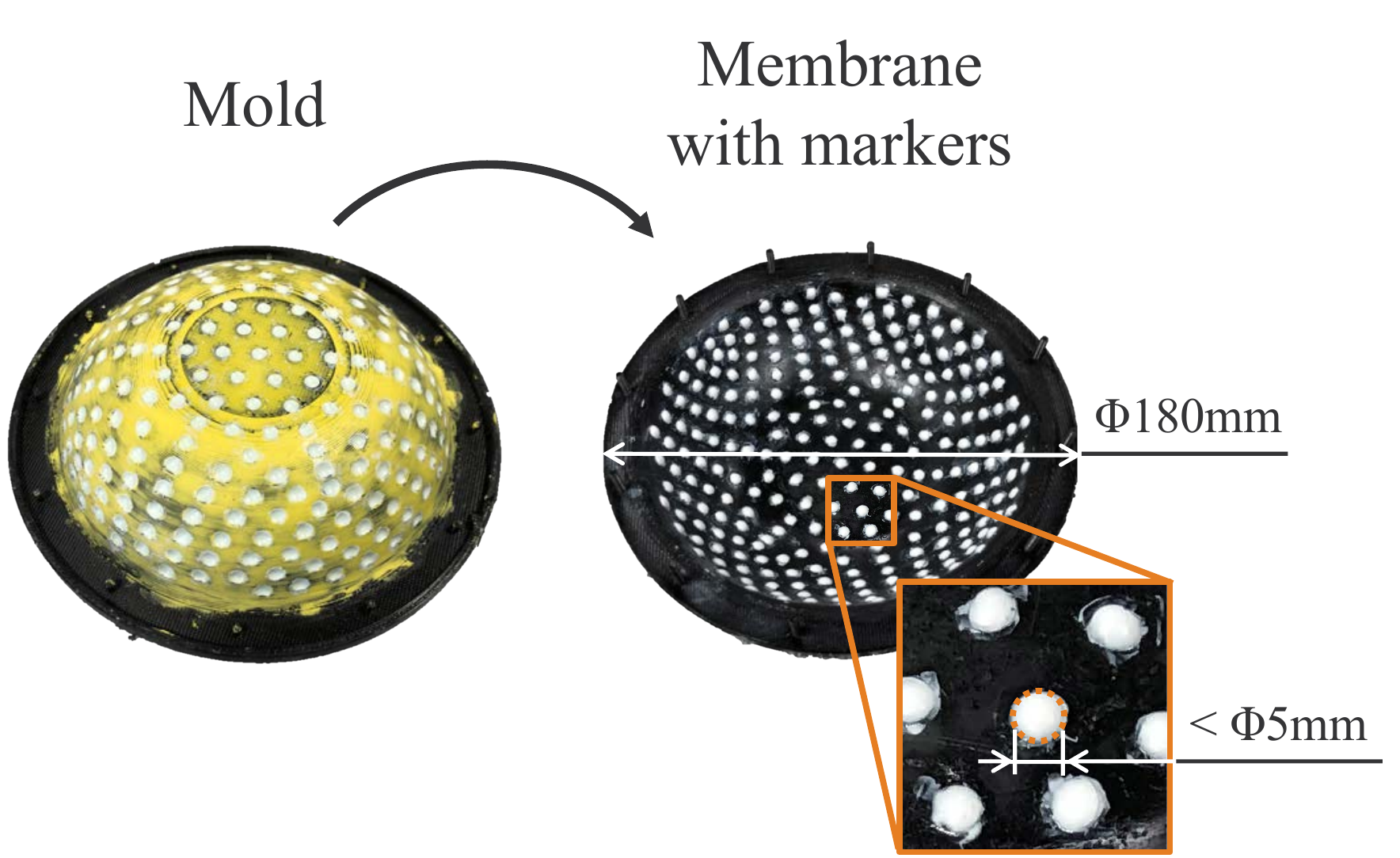}
  \caption{The mold (left) to fabricate the membrane (right) of the soft jig. The curved surface of the membrane can be formed by coating the liquid silicone material onto the rigid mold.}
  \label{fig:marker}
\end{figure}

\section{Design of soft-jig}

The soft jig shown in Fig.~\ref{fig:design}, is composed of the following four key components:

\begin{itemize}
\item A silicone membrane with markers to capture the surface deformation
\item Glass beads for jamming transition
\item A base plate cut out of transparent PET plastic
\item Two cameras to acquire the point cloud data around the target object using triangulation
\end{itemize}

The base plate size is \SI{220x220x5}{mm}, and the height of the silicone membrane is approximately \SI{30}{mm} from the base plate.
The actual soft-jig is filled with oil inside the membrane, and the markers can be seen even when the membrane is filled with beads, as shown in Fig.~\ref{fig:filling}.

\subsection{Silicone membrane}
A silicone membrane with multiple rounded convex markers was formed using the mold shown in Fig.~\ref{fig:marker}.
The thickness of the membrane is less than \SI{1}{mm} (approximately \SI{0.7}{mm}).
To form a thin membrane, we employed a method for coating liquid silicone materials (Dragon Skin FX-Pro, Smooth-On) on a mold surface instead of casting the material with several molds.
The mold had hemispherical indentations to create markers. The indentations were filled with white silicone that was colored with pigment (Silc Pig, Smooth-On) and mixed with a thickener (THI-VEX, Smooth-On).

After the white silicone hardened, we adhered the black silicone to the mold surface after filling the indentations.
Instead of using the color of the silicone itself, it was colored to make it easier to detect the marker in image processing.
The mold was printed with a 3D printer.
To easily remove the hardened silicone from the mold surface, we puttied the surface to be as smooth as possible.

\subsection{Glass beads}
In the case of large beads and a thin membrane, the shape of the bead is bumped on the surface of the membrane, which changes the contact between the membrane and the object from a surface contact to a point contact.
This causes a decrease in the coefficient of friction between the object and the membrane~\cite{paccapelo2021}.
The beads must not only be small in diameter to avoid degrading the fixation performance, but also transparent to use optical sensing.

Based on these requirements, we selected glass beads (FGB-20, Fuji Manufacturing) with a diameter of \SI{1}{mm}.

\subsection{Base plate}
The plate serves to hold a silicone membrane, optical markers, and tube connectors. The silicone membrane and tube connector were fixed by screwing, and the optical markers were fixed by inserting a pin.

Tube connectors with a filter are attached to the underside of the base plate to pump oil while preventing beads from flowing out.
Because of the small diameter of the beads, the filters need to be finer, which reduces the mass flow rate $Q$, as shown in the equation: $Q = \rho A V = \mathrm{const}$, where $\rho$ is the oil density, $A$ is the cross-section of the tube, and $V$ is the velocity.
In the production line, it is necessary to increase velocity $V$ to shorten the tact time. For this purpose, eight tube connectors were attached to the base plate.

\subsection{Camera}
Two cameras (RealSense D435, Intel) were placed approximately \SI{150}{mm} below the base plate with a baseline of \SI{30}{mm}.
Note that although this camera is a depth camera, it is used only to acquire color images with image rectification in the camera processor.

Cameras were covered with a board to remove the influence of the outside world, and an LED light maintained a constant brightness.

\section{Optical Sensing}\label{sec:optical-sensing}

\begin{figure}[t]
  \centering
  \includegraphics[width=0.7\columnwidth]{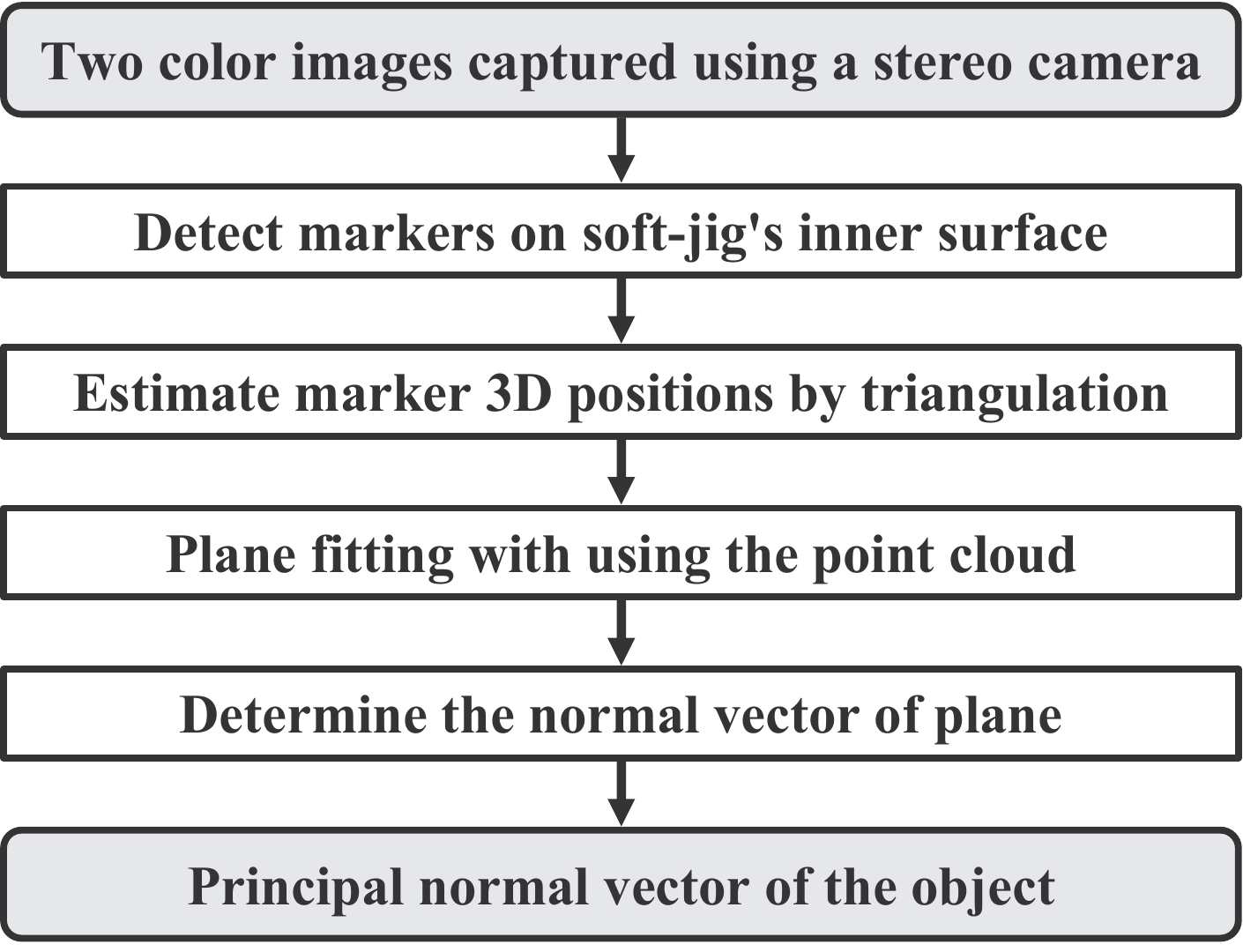}
  \caption{A flow chart of the proposed optical sensing. The inputs are two paired images, and the output is the principal normal vector of the object.}
  \label{fig:optical-sensing}
\end{figure}

\begin{figure}[t]
  \centering
  \includegraphics[width=0.6\columnwidth]{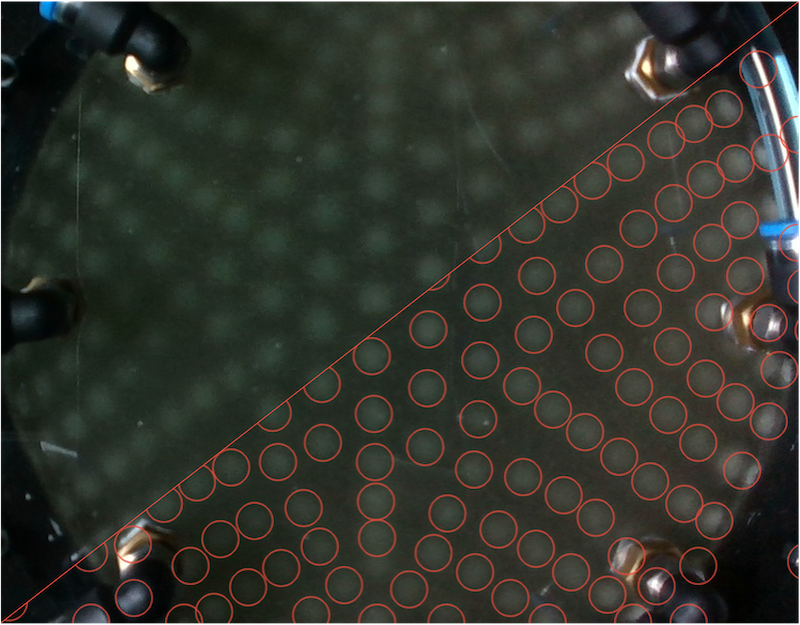}
  \caption{Appearance of markers on the inner surface of the soft jig when viewed from the camera side. The markers are blurred, making it difficult to determine their exact size; nevertheless, their positions can be captured. The red circles superimposed on the image are the result of the marker detection by LoG.}
  \label{fig:blurred}
\end{figure}

Fig.~\ref{fig:optical-sensing} shows a flowchart of the proposed optical sensing procedure. Fig.~\ref{fig:blurred} shows the markers seen through the oil.
The markers on the inner surface of the membrane are visible owing to the refractive index-tuned oil, even though the light travels through many beads to reach the membrane.
On the other hand, the refractive indices are not perfectly matched, so they appear blurred in the camera image.

The Laplacian of Gaussian (LoG) was used to accurately detect marker centers in blurred images. Using the known variation of the marker size in the image, $\sigma$ of the Gaussian filter was determined, and the center of the markers was calculated by extracting the local maximum from LoG image.

The centers of the markers detected in the left and right images were acquired with sub-pixel accuracy through least-squares fitting to a quadratic function with the center neighborhood, and converted to a point cloud by triangulation.

We used singular value decomposition (SVD) to estimate the plane of the inner surface of the soft jig along the bottom surface of the object, and it is expressed as $ax + by + cz + d = 0$ from point cloud $P(x,y,z)$.
Here, ($a, b, c$) shows the principal normal vector of the objects, and $d$ is the amount of the object pushed into the soft jig.

Note that if plane estimation using SVD is performed for all point clouds, the plane is incorrectly estimated by the markers in the areas where the object bottom surface is not in contact.
We solve this problem by using fiducial markers on the base plate of the jig, the point cloud is transformed to the jig center coordinates, and only the point cloud within the radial distance from the jig center that is less than the base area of the object is cropped.

\section{Experiments and Results}

\subsection{Principal Normal Vector Estimation}

\begin{figure}[t]
  \centering
  \includegraphics[width=0.9\columnwidth]{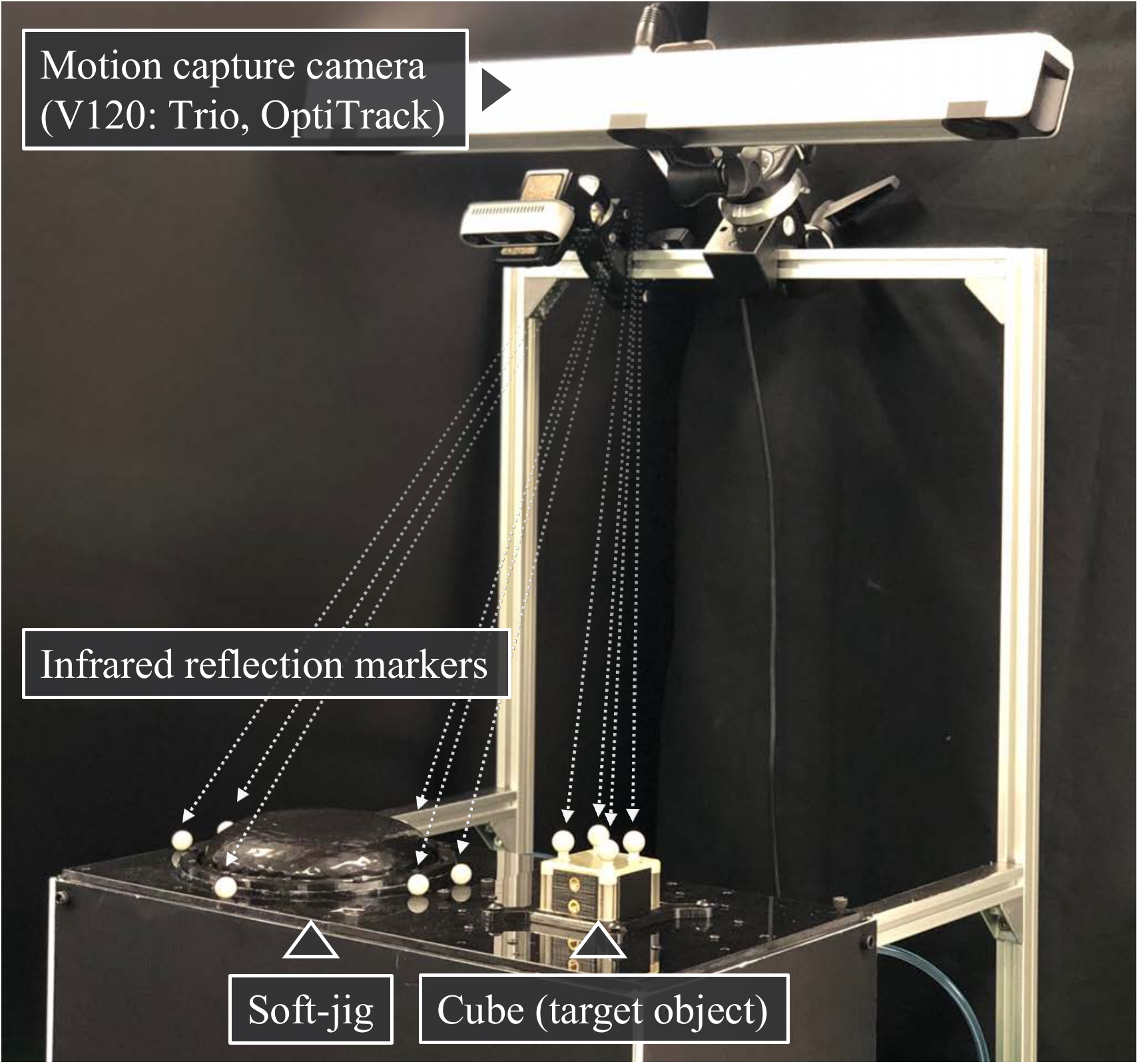}
  \caption{Setup to evaluate estimation performance of the soft jig. A motion capture camera was mounted from above to capture objects in various orientations on the jig.}
  \label{fig:mocap}
  \vspace*{10mm}
  \includegraphics[width=1.0\columnwidth]{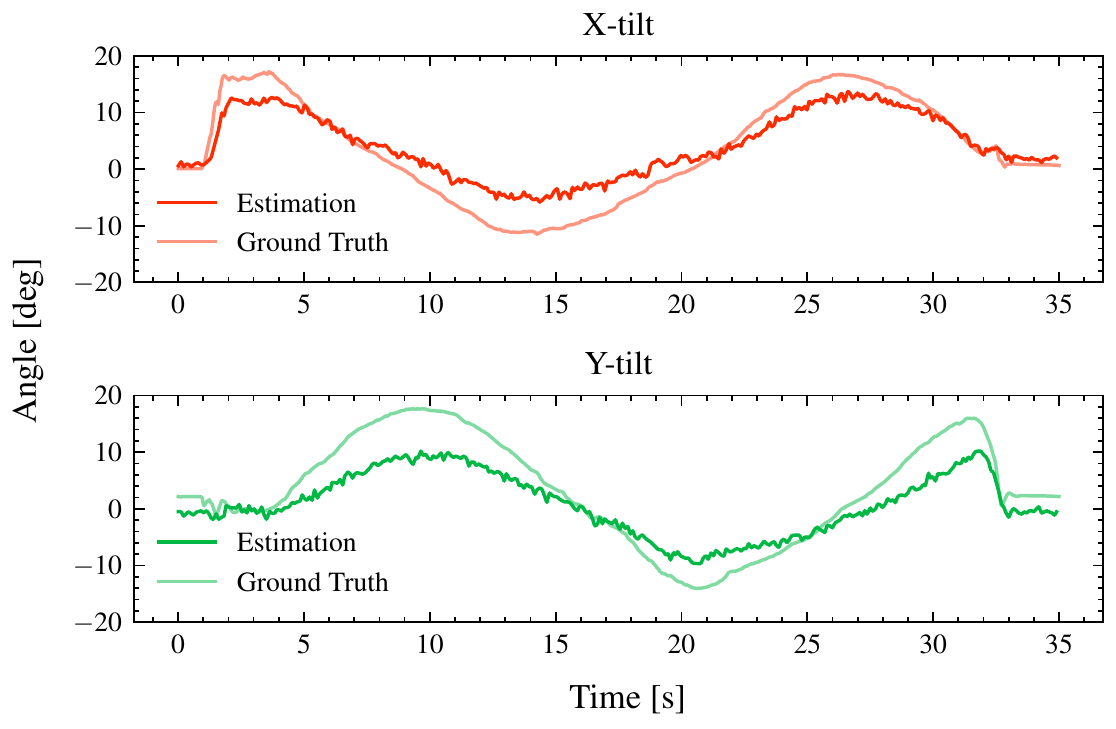}
  \caption{Performance when estimating principal normal vector of the target object (cube) using the soft jig}
  \label{fig:angle-error}
\end{figure}

We compared the difference between the ground truth data obtained by motion capture and the object's principal normal vector measured from the inside soft jig.
The six reflective markers on the jig and the four reflective markers on the cube were measured from the top using motion capture (V120: Trio, OptiTrack), as shown in the setup in Fig.~\ref{fig:mocap}.

We assume that the captured orientation of the jig is a rotation matrix $\mathbf{R}_{j}$ and translation vector $\mathbf{T}_{j}$, and captured the orientation of the target object as a rotation matrix $\mathbf{R}_{o}$ and translation vector $\mathbf{T}_{o}$.
The principal normal vector of the jig $\mathbf{n}_{d}$ in the marker coordinate system is calculated using the following equation:

\begin{eqnarray}
    \left(
    \begin{array}{cc}
        \mathbf{R}_{d} & \mathbf{T}_{d} \nonumber\\
        \mathbf{0}^T & 1
    \end{array}
    \right) &=&
    \left(
    \begin{array}{cc}
        \mathbf{R}_{j} & \mathbf{T}_{j} \\
        \mathbf{0}^T & 1
    \end{array}
    \right)^{-1}
    \left(
    \begin{array}{cc}
        \mathbf{R}_{o} & \mathbf{T}_{o} \\
        \mathbf{0}^T & 1
    \end{array}
    \right),\\
    \mathbf{n}_{d} &=& \mathbf{R}_{d} \times (0, 0, 1)^{\mathrm{T}}. \nonumber
\end{eqnarray}

The results of the projection of the normal vectors estimated from the jig $\mathbf{n}_{j}$ and the normal vectors obtained from the motion capture $\mathbf{n}_{d}$ onto the YZ and XZ planes are shown in Fig.~\ref{fig:angle-error}.
We can see that the locations of the peaks and valleys of the ground truth and the estimated values are consistent, but the absolute values of the angles are different.
This result shows that the estimated values need to be corrected by calibration in order for the robot to use these values to manipulate the object.

\begin{figure}[t]
  \centering
  \includegraphics[width=1.0\columnwidth]{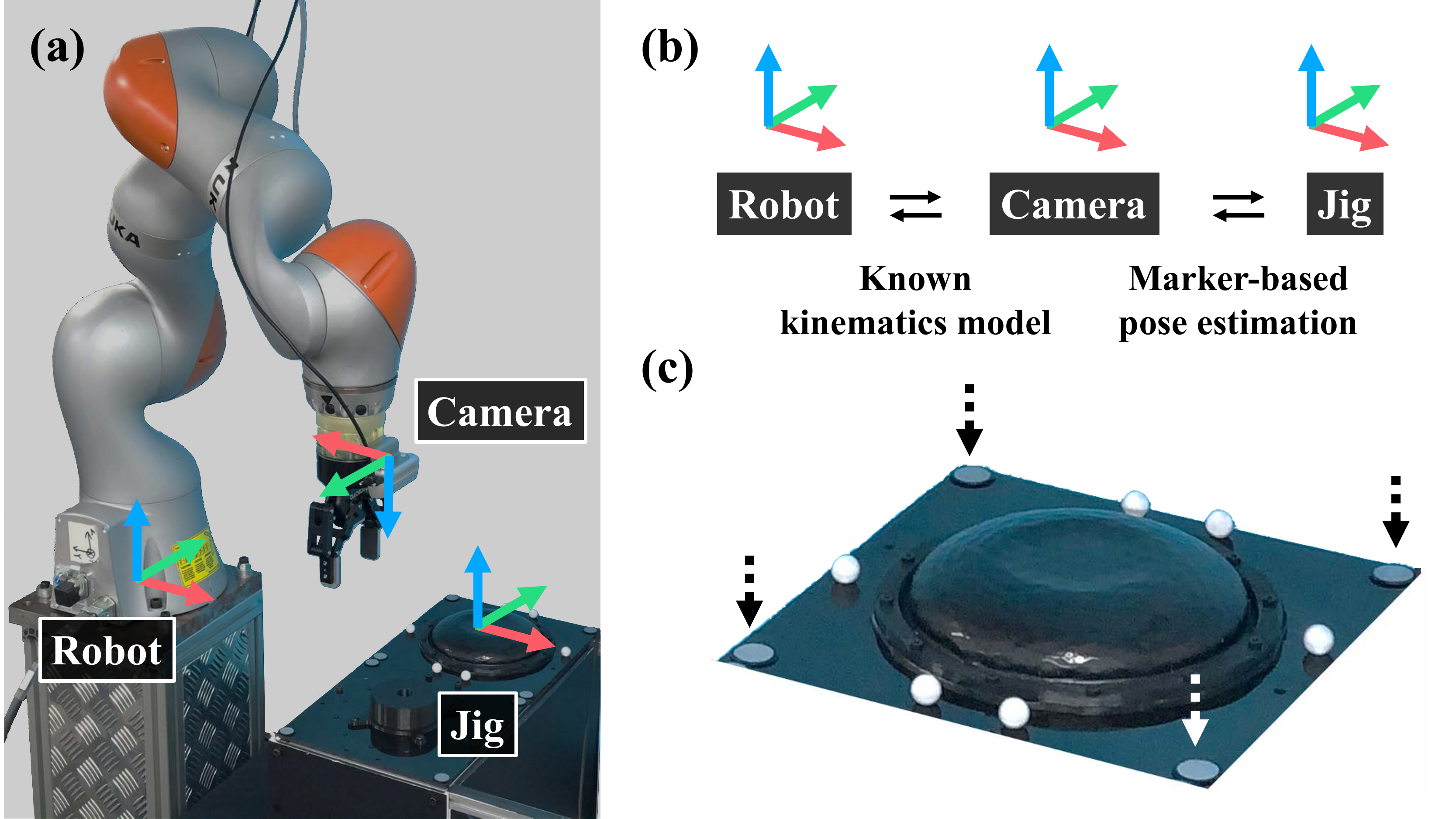}
  \caption{Setup to jig coordinate calibration. (a) A manipulator was used to push the object in a fixed orientation. (b) Transformation relations for each coordinate. The transformation from the robot to the camera was computed using the model information and kinematics. The transformation from the camera to the jig was computed using known camera parameters and marker placement. (c) Four white circular markers at the end of the dotted arrows are used to align the soft jig with the robot coordinates.
 }
  \label{fig:calib}
\end{figure}

\subsection{Jig Coordinate Calibration}

We performed calibration to convert the information acquired from the camera inside the jig into a coordinate system based on the marker attached to the base plate. For calibration, we used a manipulator (LBR iiwa 14 R820, KUKA) to precisely control the orientation of the object pressed against the jig. The robot and soft jig were arranged as shown in Fig.~\ref{fig:calib}. To transform the orientation of an object from the robot coordinate system to the jig coordinate system, a hand-eye camera and four circular markers on the jig were used for coordinate alignment using the perspective-n-point problem~\cite{lepetit2009}.

\begin{figure}[t]
  \centering
  \includegraphics[width=0.7\columnwidth]{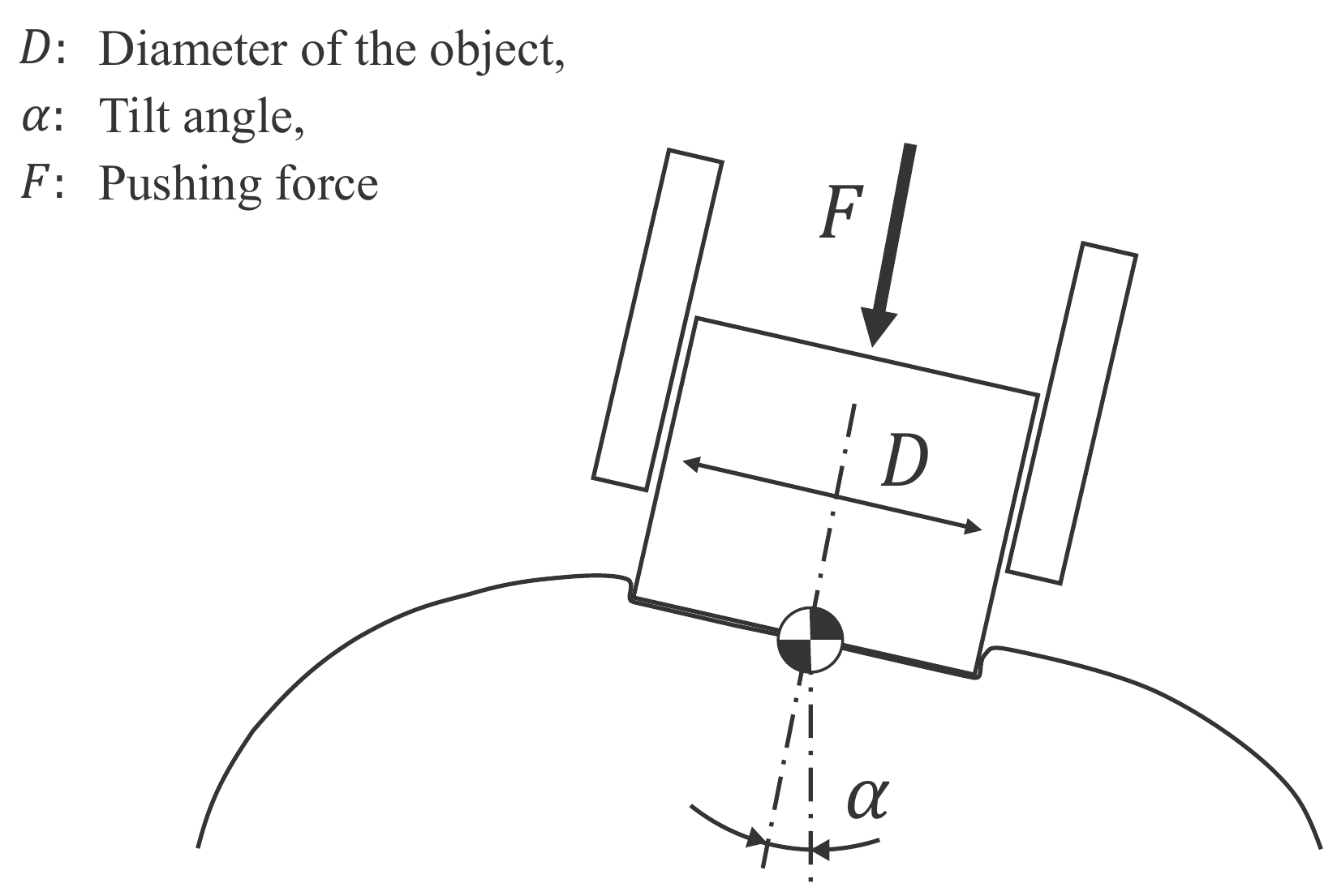}
  \caption{Schematic diagram during calibration. Data acquisition was performed by changing the size and tilt angle of the object.}
  \label{fig:push-motion}
\end{figure}

Fig.~\ref{fig:push-motion} shows a diagram of a motion of the robot to press object against the jig.
Because the orientation is estimated using the bottom surface, the estimation accuracy decreases if the bottom surface becomes smaller. The deviation between the actual and estimated angles will be large at large contact angles. To investigate the sensing abilities in terms of the object size and object orientation, we conducted estimation experiments using cylindrical objects with three different diameters $D$ tilted at different angles $\alpha$.

\begin{figure}[t]
  \centering
  \includegraphics[width=0.9\columnwidth]{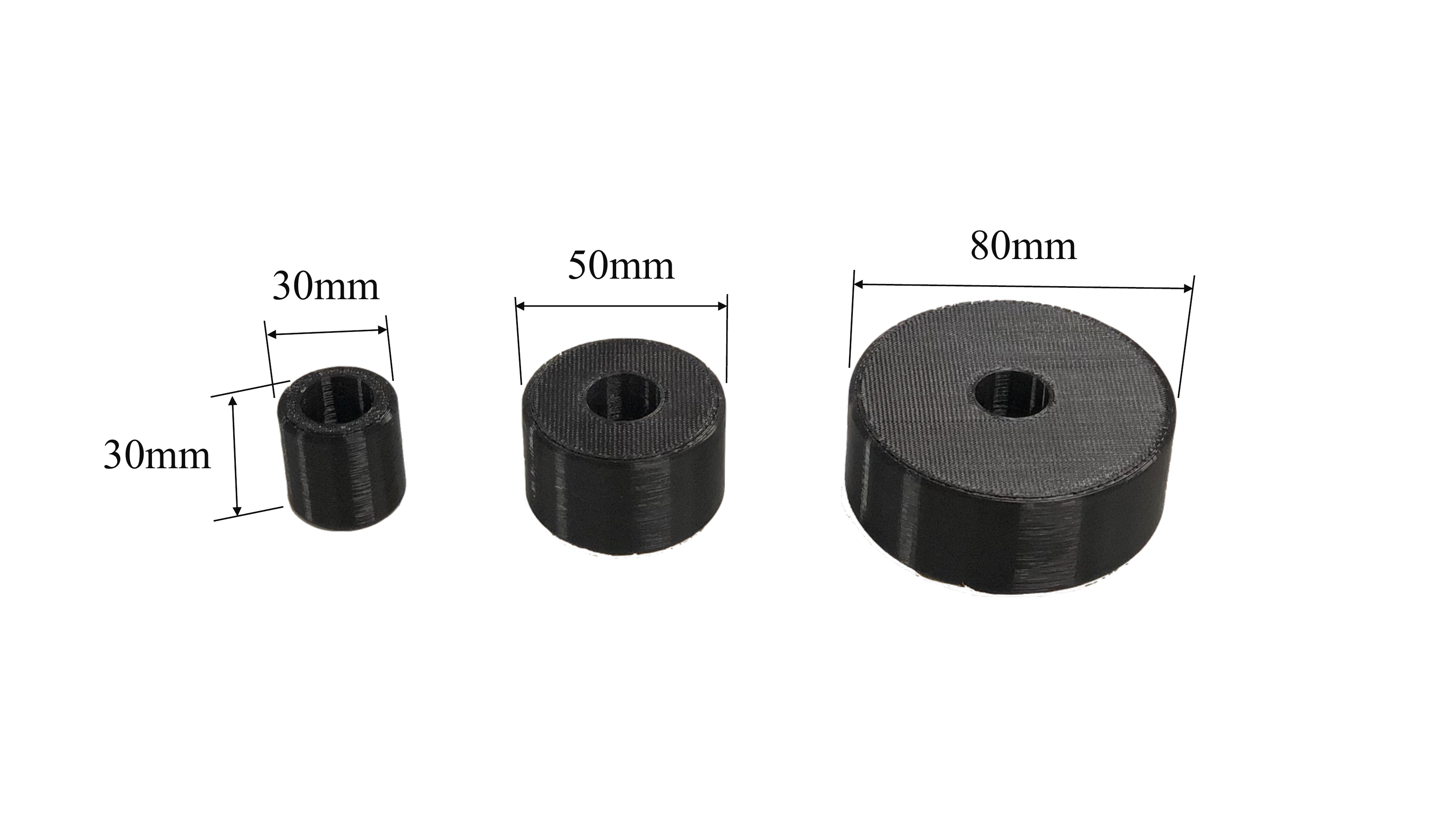}
  \vspace{-5mm}
  \caption{Cylindrical objects between \SI{30} and \SI{80}{mm} diameter. All the objects have the same height as \SI{30}{mm}.}
  \label{fig:cylindrical-objects}
\end{figure}

Fig.~\ref{fig:cylindrical-objects} shows cylindrical objects used for calibration.
The robot grasps these objects and moves to the origin within the horizontal orientation in the XY plane of the jig coordinate system using the marker recognized by the camera.
The object was then moved by the robot in the Z direction until it contacted the jig, and then it was tilted by the robot to the specified angles~$\alpha$ between 5$^{\circ}$, 10$^{\circ}$, 15$^{\circ}$, and 20$^{\circ}$ from the center of the contacting bottom. The contact with the jig was judged when the force applied to the wrist of the manipulator changed by \SI{1}{N} or more.
The sensing results from inside the jig were acquired while changing the tilt direction by 360$^{\circ}$. 

\begin{figure}[htbp]
  \centering
  \includegraphics[width=1.0\columnwidth]{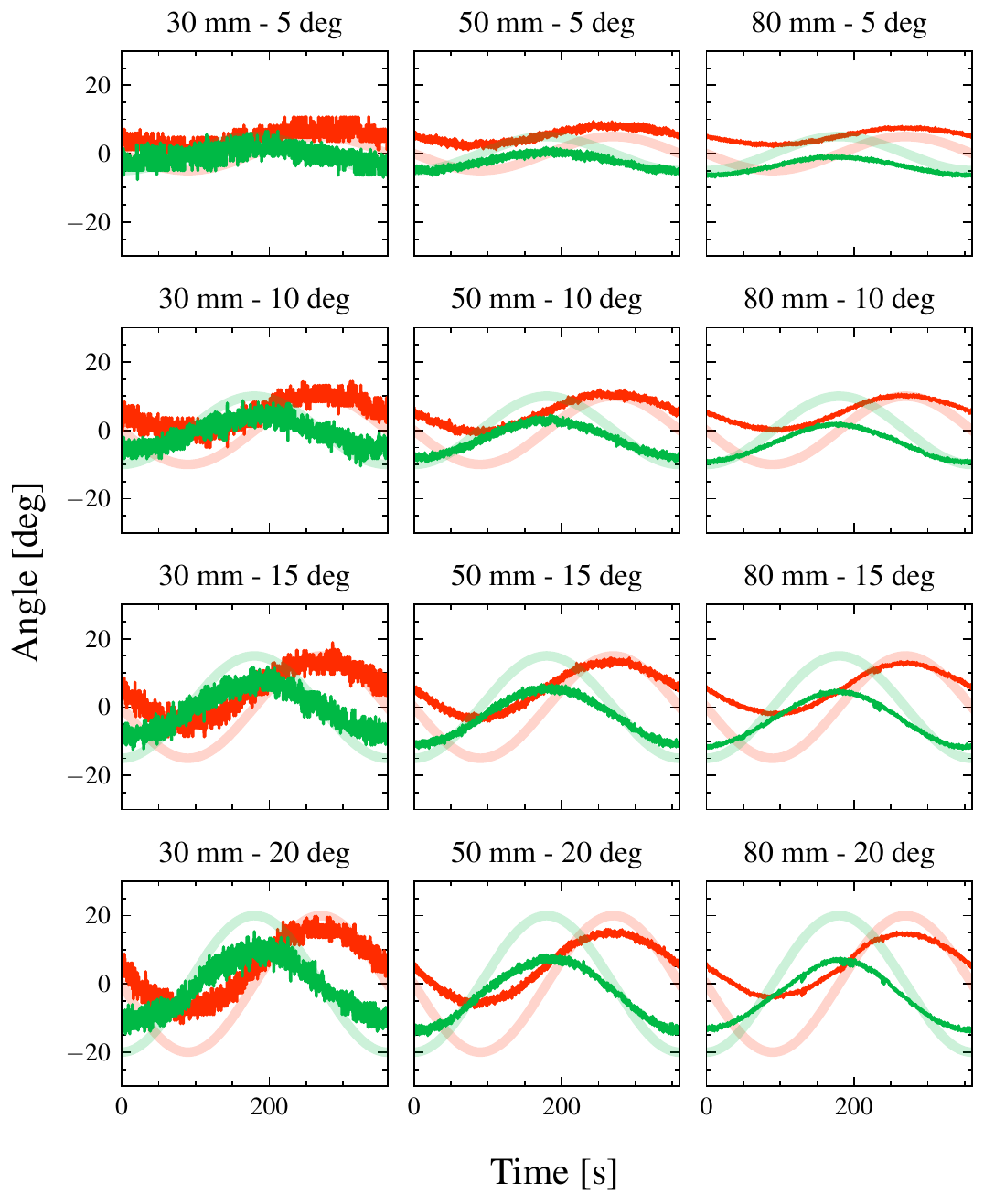}
  \caption{Tilt angles before calibration. The opaque red and green lines are the X- and Y-tilts acquired from the jig, and the translucent red and green lines are the X- and Y-tilts controlled by the robot, respectively.}
  \label{fig:tilt-angles}
  \vspace*{3mm}
  \includegraphics[width=1.0\columnwidth]{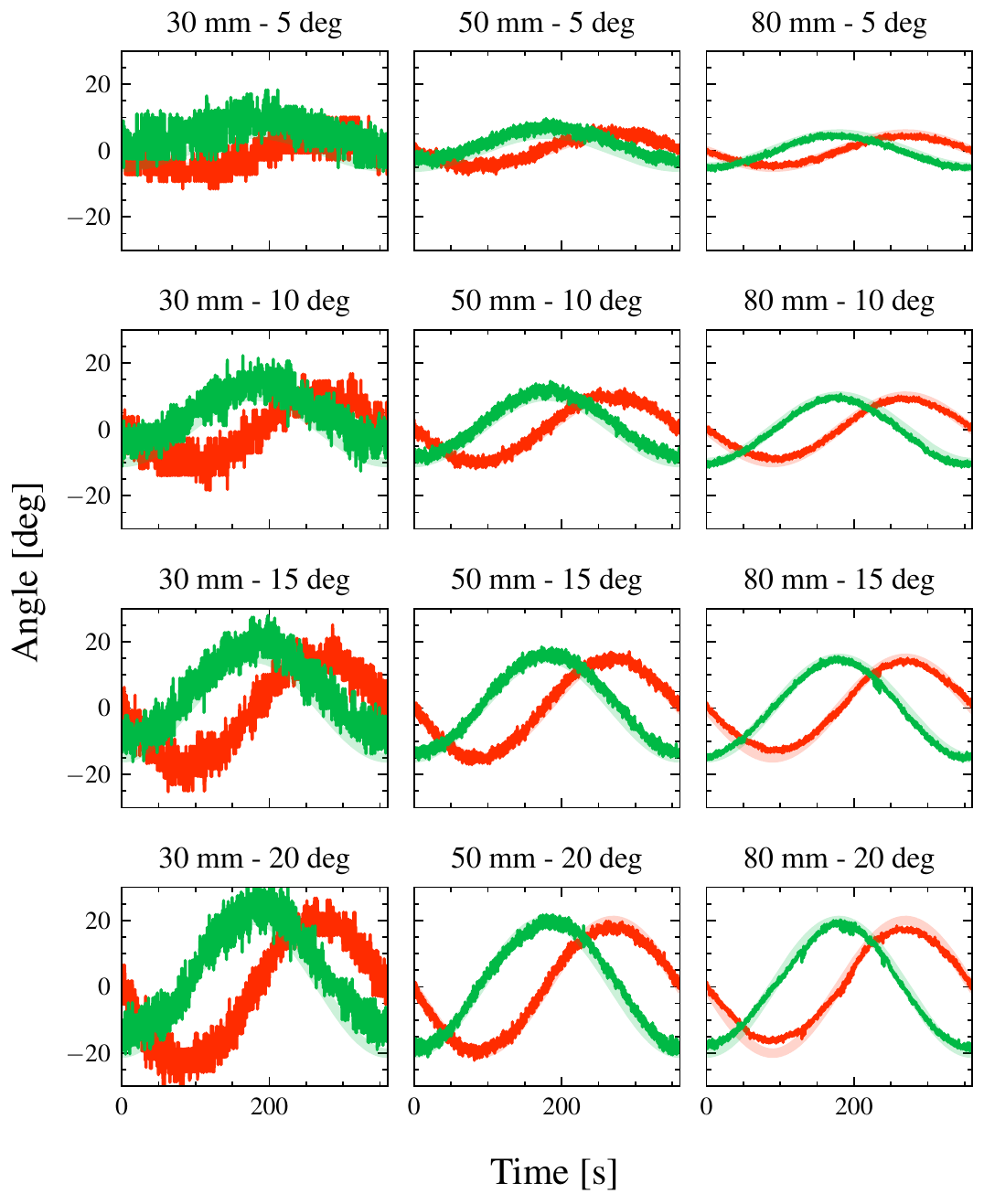}
  \caption{Tilt angles after calibration. All data were corrected using the same offsets and scales.}
  \label{fig:tilt-angles-calib}
\end{figure}

The acquired tilt angles $Ax_i$ and $Ay_i (i = 1,2,..,n)$ during calibration are shown in Fig.~\ref{fig:tilt-angles}.
$n$ is the number of samples used for the calibration.
$Ax_i$ and $Ay_i$ were calculated by projecting the normal vector onto the YZ and XZ planes and converting them into angles.
We calculated three parameters, offsets $O_x, O_y$, and scale $S$, as calibration results to obtain the correct value of the acquired tilt angle.
$Ax_i$ and $Ay_i$ can be regarded as a pure sine wave in the YZ and XZ planes because the tilt direction is changed by 360$^{\circ}$ while keeping the angle $\alpha$ constant in the radial direction by the robot.
Therefore, the offset $O_x, O_y$, and scale $S$ can be calculated using the following equations:

\begin{equation}
    O_x = \frac{1}{n} \sum_{i=1}^n Ax_i,\ 
    O_y = \frac{1}{n} \sum_{i=1}^n Ay_i,\ 
    S = \frac{2 \times {\alpha}}{\sqrt{2} \times (\mathbf{\sigma_x}+\mathbf{\sigma_y})} \nonumber,
\end{equation}

where:
\begin{equation}
    \mathbf{\sigma_x} = \sqrt{\frac{1}{n} \sum_{i=1}^n (Ax_i - O_x)^2}, 
    \mathbf{\sigma_y} = \sqrt{\frac{1}{n} \sum_{i=1}^n (Ay_i - O_y)^2} \nonumber.
\end{equation}

\begin{figure}[t]
  \centering
  \includegraphics[width=1.0\columnwidth]{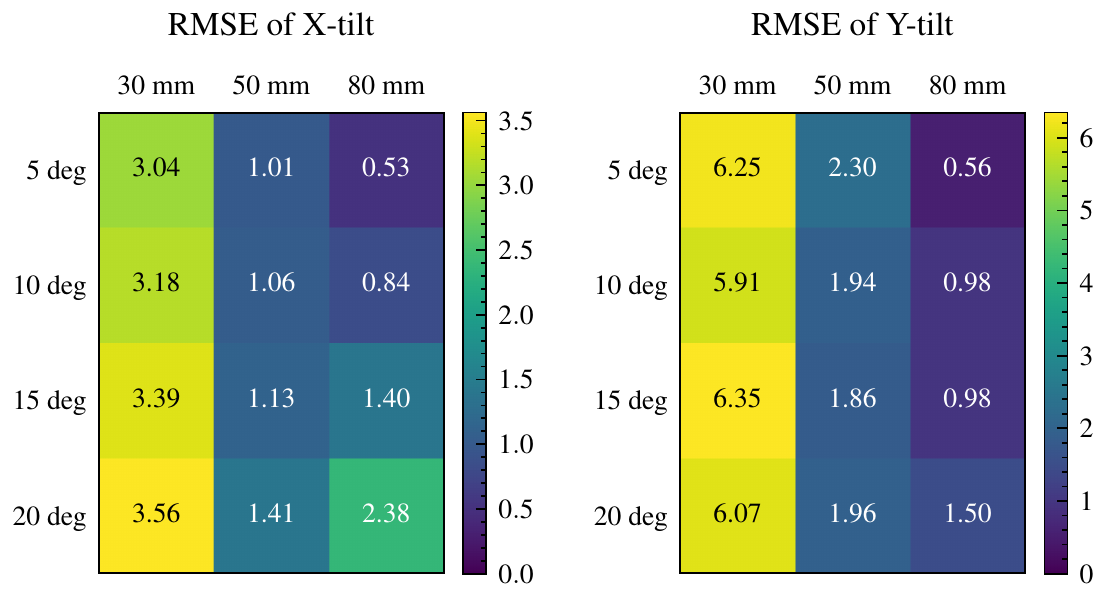}
  \caption{The RMSE of tilt angle after calibration arranged in grids. These grids correspond to the arrangement shown in Figs. 12 and 13.}
  \label{fig:rms-grid}
\end{figure}

The tilt angles calibrated using these parameters are presented in Fig.~\ref{fig:tilt-angles-calib}. The root-mean-square error (RMSE) of the difference between the tilt angles and the angles controlled by the robot is shown in Fig.~\ref{fig:rms-grid}.

When the diameter of the object was small, the number of markers that could be referenced was small; the result of the normal estimation became noisy. If the diameter of the bottom surface was larger than 50 mm, the noise would have been reduced, and the RMSE was stabilized.
In addition, the values in the lower right corner of the RMSE table are larger. This is probably due to the fact that when a large object is tilted and pressed against the jig at a large angle, part of the bottom surface of the object lifts up and becomes non-contact with the jig.

\section{Discussion}

Our proposed soft jig suffers from some limitations in terms of material selection.
The blurred image is caused by the fact that the refractive indices of the beads and liquid do not exactly match, and also by the fact that the beads are not completely clear.
In this study, we used inexpensive glass beads for polishing to fill 500 g of beads, but more transparent materials, such as acrylic beads, would be more suitable.

The current issue is that a camera from the outside world is required for position estimation.
This is because it is difficult to estimate the position of the \SI{1}{mm} order from the tactile information because of the insufficient density of marker placement.
However, we believe that by introducing a mechanism that can change the color of the membrane to visible or invisible as in a previous study~\cite{hogan2021}, we will be able to estimate the uncertainty of the position as well.

\section{Conclusion}

We proposed a soft jig with a sensing function for an assembly system, a method for estimating a principal normal vector using tactile information, and a calibration method using a real robot.
Conventional flexible jigs do not consider the uncertainty caused by flexibility, but the proposed jig with a sensing function can simultaneously fix and estimate the object's orientation. The principal normal vector can be used to refine the object's orientation acquired from an external camera during object manipulation, even when the object is occluded.

\bibliographystyle{IEEEtran}
\bibliography{bibliography/references}

\end{document}